\documentclass[letterpaper, 10 pt, conference]{ieeeconf}  

\IEEEoverridecommandlockouts                              

\title{\LARGE \bf
A Data-Driven Approach for Autonomous Motion Planning and Control in Off-Road Driving Scenarios
}

\usepackage{blindtext}
\usepackage{graphicx}
\usepackage[utf8]{inputenc}
\usepackage{graphics} 
\usepackage{epsfig} 
\usepackage{mathptmx} 
\usepackage{amsmath} 
\usepackage{amssymb}  
\usepackage{cite}
\usepackage{multicol}
\usepackage{mathtools, cuted}
\usepackage[cmintegrals]{newtxmath}
\usepackage{epstopdf}
\usepackage{graphics} 
\usepackage{epsfig} 

\usepackage{mathptmx} 
\usepackage{amsmath} 
\usepackage{amssymb}  
\usepackage{cite}
\usepackage{multicol}
\usepackage{mathtools, cuted}
\usepackage[cmintegrals]{newtxmath}
\usepackage{mathtools}

\usepackage{mathptmx}
\usepackage{helvet}
\usepackage{courier}
\usepackage{subfigure}
\usepackage{type1cm}
\usepackage{titlesec}
\usepackage{epsfig} 
\usepackage{mathptmx} 
\usepackage{amsmath} 
\usepackage{amssymb}  
\usepackage{cite}
\usepackage{multicol}
\usepackage{mathtools, cuted}
\usepackage[cmintegrals]{newtxmath}
\usepackage{algpseudocode}
\usepackage{makeidx}         
\usepackage{algorithm}
\usepackage{algpseudocode}

\usepackage{multicol}        
\usepackage[bottom]{footmisc}
\usepackage{caption}
\usepackage{nomencl}
\usepackage[usenames, dvipsnames]{color}

\usepackage{multicol}        
\usepackage[bottom]{footmisc}
\usepackage{caption}
\usepackage{titlesec}
\usepackage{nomencl}
\usepackage[usenames, dvipsnames]{color}
\usepackage{multirow}

\usepackage{amsthm}
 
\theoremstyle{definition}
\newtheorem{definition}{Definition}

\theoremstyle{theorem}

\theoremstyle{remark}
\newtheorem{remark}{Remark}

 \title{\LARGE \bf
 ``Sandwich'' Approach for Motion Planning  and Control
}

\begin{document}

\author{Hossein Rastgoftar and Mohamadreza Ramezani
\thanks{Authors are with the Aerospace and Mechanical Engineering Department,
 University of Arizona, Tucson,
AZ, 85719 USA e-mail: \{hrastgoftar, ramezani\}@arizona.edu.}
}

\maketitle
\thispagestyle{empty}
\pagestyle{empty}

\begin{abstract}
This paper develops a new approach for robot motion planning and control in obstacle-laden environments that is inspired by fundamentals of fluid mechanics. For motion planning, we propose a novel transformation between  ``motion space'', with arbitrary obstacles of random sizes and shapes, and an obstacle-free  ``planning space'' with geodesically-varying distances and  constrained transitions. We then obtain robot's desired trajectory by A* searching over a uniform grid distributed over the planning space. We show that  implementing the A* search over the planning space can generate  shorter paths when compared to the existing A* searching over the motion space. For trajectory tracking, we propose an MPC-based trajectory tracking control, with linear equality and inequality safety constraints, enforcing the safety requirements of planning and control. 

\end{abstract}

\section{INTRODUCTION}
Obstacle-laden environments  can be majorly constrained by  stationary objects that must be taken into account when planning an area's use even by a single robot. Indeed, this is  computationally inefficient.  Motion planning becomes even more computationally inefficient a robot multiple times or multiple robot  simultaneously use the motion space since assuring collision avoidance requires counting obstacles for every robot when  a single robot uses the motion space for multiple times, or multiple robots  simultaneously use the motion space, since ensuring collision avoidance will need to count all obstacles for every single robot any time. To overcome this problem, we propose to convert an obstacle-laden motion space into an obstacle-free planning space with constrained transitions and ensure collision avoidance by planning the robot motion in the planning space. 

\subsection{Related Work}\label{Related Work}
Control Barrier Function (CBF) approach has been applied  for safety assurances and collision avoidance \cite{aaroncaltech2, aaroncaltech1, DimitraP1}. More specifically, CBF uses system dynamics to define an admissible region in the agent's workspace\cite{chengCBFdynms}, and the agent's control inputs are then calculated to ensure that the robot's state remains within this region at all times\cite{zhangCBF_final, IRCBF_final}. Model-predictive control with the use of control barrier function is a widely used method in safe path tracking applications\cite{safetyandMPC_marvi2019}, i.e., unmanned aerial vehicles\cite{dynamicmotion2021motion}.

 Inspired by analytical solution for ideal fluid flow over multiple cylinders \cite{crowdy2006analytical, crowdy2016uniform}, we recently proposed to model
 UAS coordination as ideal fluid flow and  provided an  analytical approach to obtain potential and stream fields of the flow field for two-dimensional (2D) motion planning \cite{rastgoftar2019physics}, where we generated the \textit{potential function} $\phi(x,y)$ and \textit{stream function} $\psi(x,y)$ by   combining uniform and double flow patterns. Indeed, this proposed solution is highly resilient to pop-up failure or unpredicted (and predicted) obstacles that are safely wrapped if UASs follow the streamlines.  
 
The analytical solution's main drawback, however, is that the geometry (shape) and  the size of the domain enclosing obstacles cannot be fully controlled. As a result, to ensure collision avoidance, we must overestimate the geometry of the domains enclosing obstacles
, which makes the motion  planning excessively conservative. To resolve this issue of analytical solution, we employed numerical methods \cite{10180409} to  obtain \textbf{only} the stream function $\psi(x,y)$ for 2D coordination of UASs in urban environments where arbitrary-sized and -located buildings were safely encircled by streamlines. However, it is challenging  to solve both PDEs \eqref{phi} and \eqref{phi} for an random distribution of obstacles  of arbitrary sizes in order to obtain $\phi(x,y)$ and $\psi(x,y)$ and establish a nonsingular transformation between $(x,y)$ and $(\phi,\psi)$ at every non-obstacle location. 

\begin{figure}[htbp]
\centering
\includegraphics[width=3.3in]{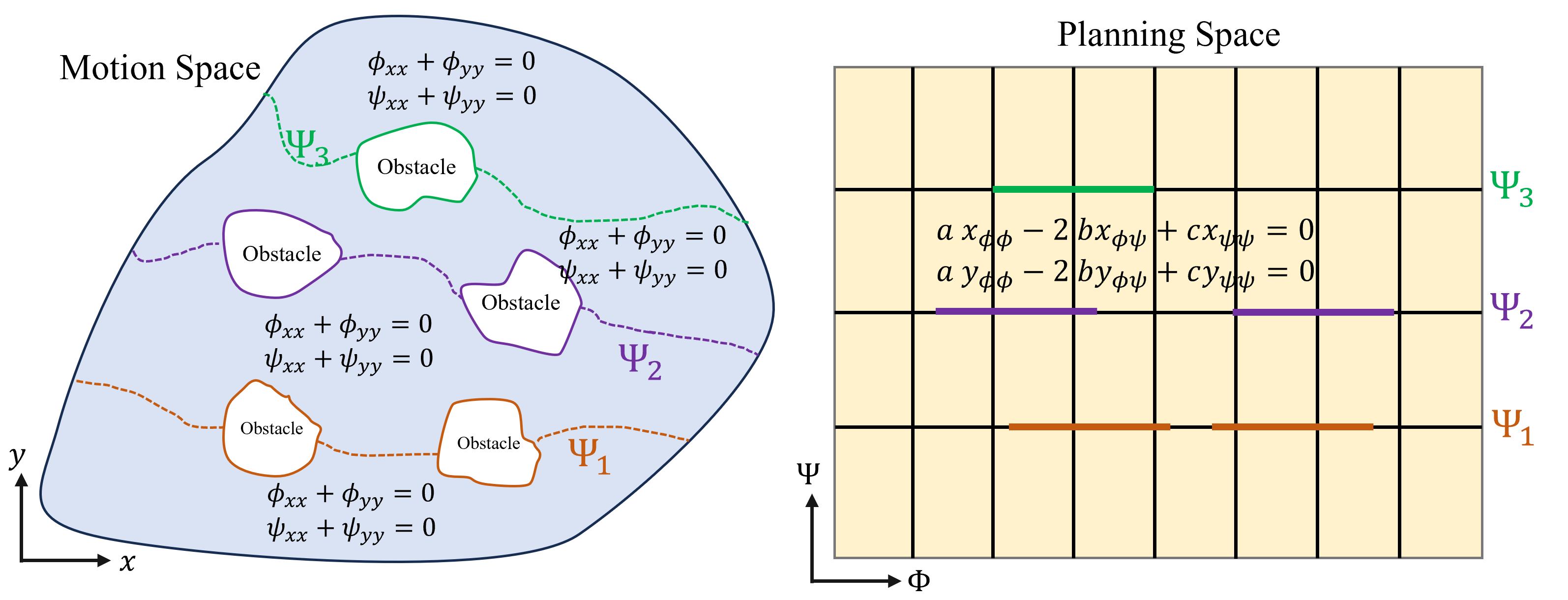}
 \vspace{-0.3cm}
\caption{Transforming the ``motion'' space (left) to the ``planning'' space (right).}
\label{MotionPlanningScheme-V2}
\end{figure}
\subsection{Contributions and Paper Outline}
We propose a novel approach for safe and collision-less motion planning, wherein each obstacle is treated as a ``rigid'' body whose boundary is determined by two streamlines, having the same steam value and enclosing (or sandwiching) the obstacle. Inspired by the mesh generation techniques used for computational fluid dynamics applications, we will overcome the issue highlighted in Section \ref{Related Work}  by (i) dividing the motion space into multiple navigable spaces, (ii) classifying obstacles into a finite number of groups, and (iii) sandwiching each obstacle group by two adjacent navigable channels (See Fig. \ref{MotionPlanningScheme-V2}). As a result, we can establish a non-singular transformation between $x-y$ and   $\phi-\psi$ plane with forbidden crossing over finite number of horizontal line segments in the planning space. Compared to the available motion planning methods, our approach offers the following contributions:
\begin{itemize}
    \item We obtain the optimal path with a shorter length by A* searching over the planning space instead of motion space.    
    \item We consider obstacles  as rigid-bodies with boundaries that cannot be trespassed since obstacles are excluded when we map the motion space to the planning space.
    \item The optimal path minimizing travel distance between the initial location and target destination is obtained by searching over a uniform grid with geodesically varying distances and constrained transition.
\end{itemize}
We apply model predictive control (MPC) for the safety-verified  trajectory tracking control where safety constraints are defined as linear inequalities.


This paper is organized as follows: Problem Statement is given in Section \ref{PROBLEM STATEMENT}. The motion planning approach  is presented as Spatial and Temporal planning problems  in Sections  \ref{Spatial Planning} and \ref{Temporal Planning}, respectively. Safety-verified MPC-based trajectory tracking is formulated in Section  \ref{Trajectory Tracking Control}. Simulation results are presented in Section \ref{Results} and followed by concluding remarks in Section \ref{Conclusion}.

\section{Problem Statement}\label{PROBLEM STATEMENT}
We consider robot motion planning in an obstacle-laden motion space and decompose it into spatial and temporal planning problems.
Section \ref{Spatial Planning} proposes to exclude obstacles by transforming the ``motion'' space into the ``planning'' space,  and ensuring  collision avoidance by robot motion planning in the planning space. The temporal planning is achieved through A* searching over a uniform grid distributed over the planning space as discussed in  Section \ref{Temporal Planning}.


For the trajectory tracking control, without loss of generality, we consider  quadcopter as the robot navigating in an obstacle-laden environment and  the dynamics developed in Refs. \cite{rastgoftar2021safe, el2023quadcopter} to model its motion. Then, the trajectory tracking control is formulated as MPC in Section \ref{Trajectory Tracking Control}.



\begin{remark}
    This paper assumes that the motion space is two dimensional where we use $x$ and $y$ to denote position components. For $3$-D motion planning, we can decompose the motion space into parallel horizontal floors where $x$ and $y$ can be used to specify position in every floor.
\end{remark}

\section{Spatial Planning}\label{Spatial Planning}
We  use $\mathcal{P}$ to denote the motion space, and  decompose it into  \textit{navigable} subspace $\mathcal{S}$ and \textit{obstacle} subspace $\mathcal{R}$ ($\mathcal{P}=\mathcal{S}\bigcup\mathcal{R}$). We apply the ideal fluid flow concept to establish a nonsingular mapping between navigable subspace $\mathcal{S}$,  with position coordinates $x$ and $y$, and the planning space,  with position coordinates $\phi$ and $\psi$, where we $\mathcal{C}$ to refer to the planning space. Note that $\phi\left(x,y\right)$ and $\psi\left(x,y\right)$ are the potential and stream fields over the motion space, and maintain the following properties:
\\
\textbf{Property 1:} Every obstacle in $\mathcal{P}$ is wrapped by a $\psi$-constant manifold which is a steamline in the motion space.
\\
\textbf{Property 2:} Potential lines $\phi$-constant and streamlines $\psi$-constant are all perpendicular at any crossing  point in $\mathcal{S}$.
Property 1 is achieved, when stream ($\psi$) values are constant along the outer boundary of every obstacle  \cite{rastgoftar2019physics}.
By achieving Property 2, both $\phi$ and $\psi$, defined over the physical domain, satisfy the Laplace partial differential equation (PDE):
\vspace{-0.25cm}
\begin{subequations}
    \begin{equation}\label{phi}
        \phi_{xx}+\phi_{yy}=0,
\end{equation}
    \begin{equation}\label{psi}
        \psi_{xx}+\psi_{yy}=0.
\end{equation}
\end{subequations}
As mentioned in Section \ref{Related Work}, obtaining $\phi$ and $\psi$ is challenging by solving Laplace PDEs \eqref{phi} and \eqref{psi}, when obstacles of random sizes and shapes are arbitrarily distributed over the motion space. We propose to define $x$ and  $y$ as functions of independent variables  $\phi$ and $\psi$, and then use the chain rules to obtain derivatives of dependent variables $x$ and  $y$ with respect to $\phi$ and $\psi$. This establishes a mapping between the navigable space $\mathcal{S}\subset \mathcal{P}$ and the planning space, and converts the Laplace PDEs \eqref{phi} and \eqref{psi}  to the following PDEs \cite{hoffmann1993computational}:
\vspace{-0.25cm}
 \begin{subequations}\label{ellipticpde}
     \begin{equation}\label{xpp}
         a\left(\phi,\psi\right) x_{\phi \phi}-2 b\left(\phi,\psi\right) x_{\phi \psi}+c\left(\phi,\psi\right) x_{\psi \psi}  =0,
     \end{equation}
     \begin{equation}\label{ypp}
         a\left(\phi,\psi\right) y_{\phi \phi}-2 b\left(\phi,\psi\right) y_{\phi \psi}+c\left(\phi,\psi\right) y_{\psi \psi}  =0,
     \end{equation}
 \end{subequations}
where $a\left(\phi,\psi\right) =x_\psi^2+y_\psi^2$, $b\left(\phi,\psi\right) =x_{\phi} x_\psi+y_{\phi} y_\psi$, and $c\left(\phi,\psi\right) =x_{\phi}^2+y_{\phi}^2$.


We use finite difference approach to solve the elliptic PDEs  \eqref{xpp} and \eqref{ypp}. To this end, we first provide a solution for dividing $\mathcal{S}$ into multiple obstacle-free navigable subspaces in  Section \ref{Conversion of the Physical Domain to the Computational Domain}. We then specify boundary conditions to determine $x\left(\phi,\psi\right)$ and  $y\left(\phi,\psi\right)$ over the planning space, by solving governing PDEs \eqref{xpp} and \eqref{ypp}. 

\begin{figure}[htbp]
\centering
\includegraphics[width=3.3in]{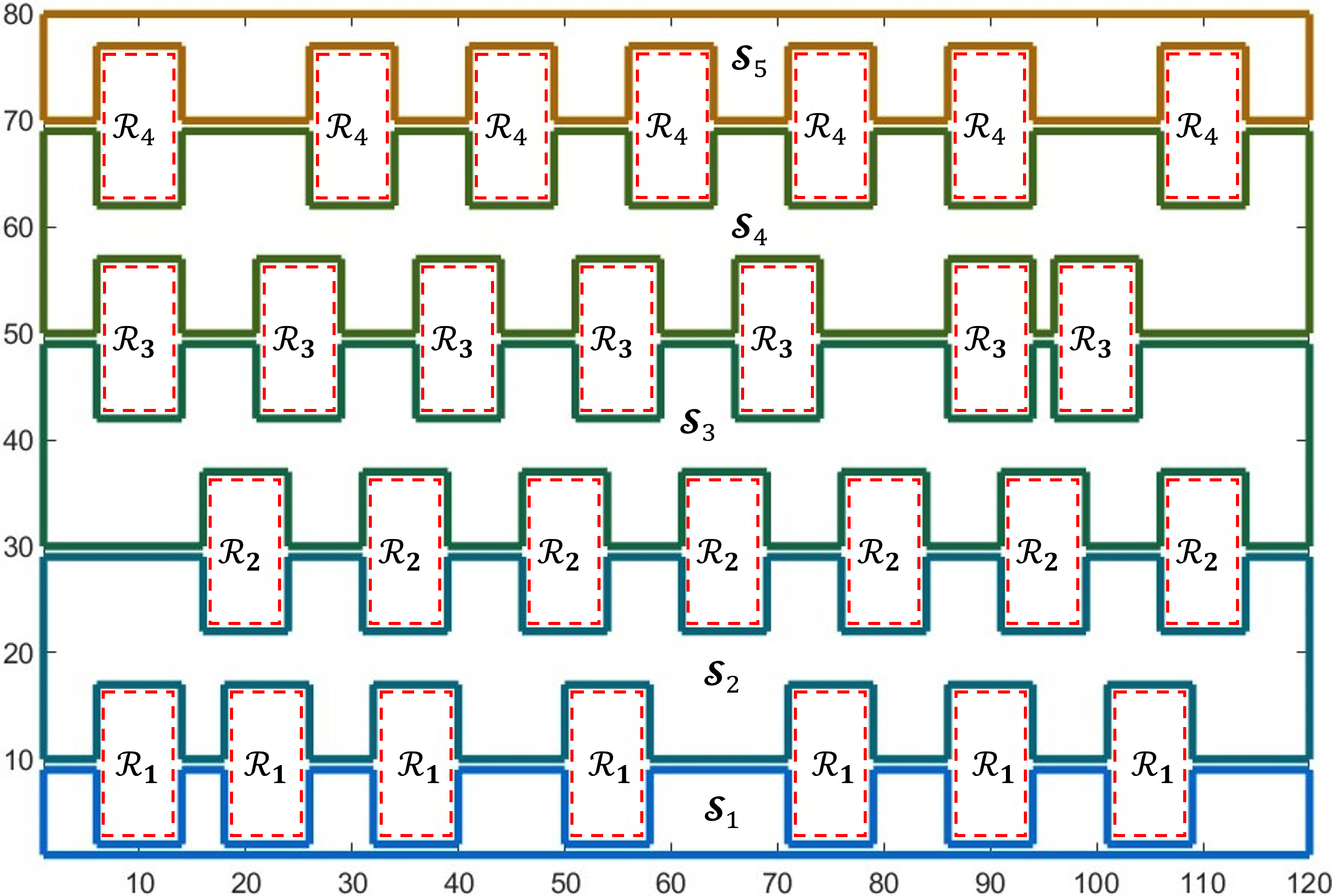}
\vspace{-0.25cm}
\caption{Dividing an example navigable space $\mathcal{S}$ into five navigable channels $\mathcal{S}_1$ through $\mathcal{S}_5$.}
\label{ExampleMotionSpace}
\end{figure}

\subsection{Motion Space Decomposition and Obstacle Sandwiching}\label{Conversion of the Physical Domain to the Computational Domain}
We consider a motion space consisting of $m_o$ obstacles, where each obstacle is a compact zone in $\mathcal{P}$, and  obstacles are identified by set $\mathcal{O}=\left\{1,\cdots,m_o\right\}$. We express $\mathcal{R}$ as
\vspace{-0.25cm}
\begin{equation}
    \mathcal{R}=\bigcup_{j=1}^{p-1}\mathcal{R}_j
\end{equation}
where $\mathcal{R}_j$ is a group of obstacles with the same $\psi$ values on their external boundaries. For example, obstacles shown in the motion space shown of Fig. \ref{MotionPlanningScheme-V2} is categorized into three groups $\mathcal{R}_1$, $\mathcal{R}_2$, and $\mathcal{R}_3$, where obstacles of $\mathcal{R}_j$ all wrap by two streamlines having the same value $\Psi_j$, for $j=1,2,3$.

\begin{figure}[htbp]
\centering
\includegraphics[width=3.3in]{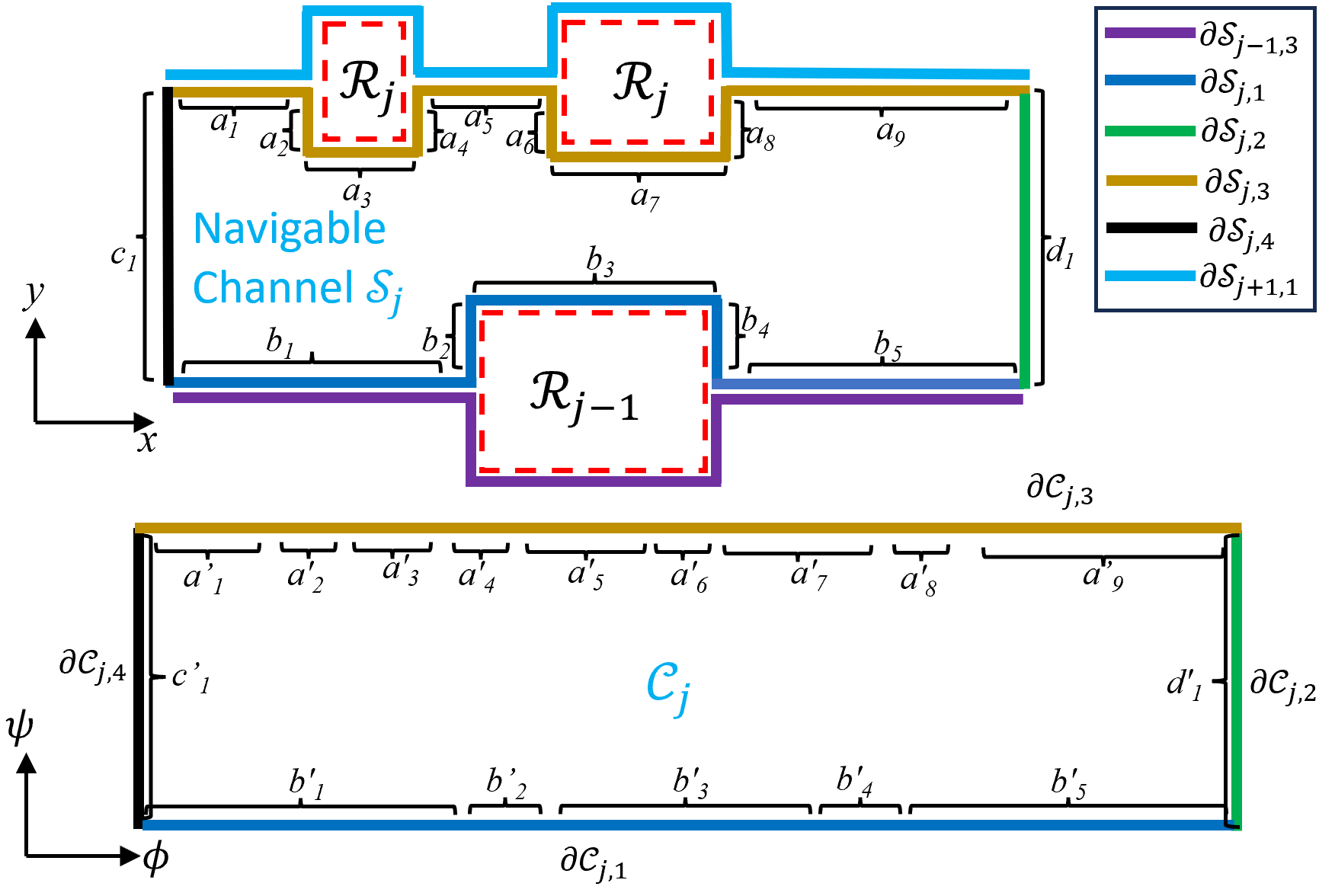}
\vspace{-0.25cm}
\caption{Mapping between the navigable channel $\mathcal{S}_j$ and $j$-th rectangle $\mathcal{C}_j$ in the planning space.}
\label{MappingSchematicChannel}
\end{figure}

Given $\mathcal{R}$, the  navigable zone $\mathcal{S}=\mathcal{P}\setminus \mathcal{R}$ is expressed as 
\vspace{-0.3cm}
 \begin{equation}
        \mathcal{S}=\bigcup_{j=1}^{p}\mathcal{S}_j,
    \end{equation}
where $\mathcal{S}_j$ is called the \textit{$j$-th navigable channel}. Note that all obstacle zones defined by $\mathcal{R}_j$ are sandwiched by navigable channels $\mathcal{S}_j$ and $\mathcal{S}_{j+1}$. For better clarification, we consider the rectangular motion space shown in Fig. \ref{ExampleMotionSpace} with  obstacles categorized as $\mathcal{R}_2$ through $\mathcal{R}_5$,  $\mathcal{S}$ is decomposed into five navigable channels ($p=5$); and $\mathcal{R}_{j+1}$ is sandwiched by navigable channels $\mathcal{S}_j$ and $\mathcal{S}_{j+1}$ for $j=1,\cdots,p-1$.  We use $\partial \mathcal{S}_{j,1}$, $\partial \mathcal{S}_{j,2}$, $\partial \mathcal{S}_{j,3}$, and $\partial \mathcal{S}_{j,4}$ to define the bottom, right, top, and left boundaries of $\mathcal{S}_j$, respectively (See Fig. \ref{MappingSchematicChannel}).  

\subsection{Planning Space Decomposition}\label{BoundaryConditionsSpecification}
The planning space is a rectangle defined by 
\begin{equation}
    \mathcal{C}=\left\{\left(\phi,\psi\right):\phi\in \left[\phi_{min},\phi_{max}\right],~\psi\in \left[{\Psi}_1,{\Psi}_{p+1}\right]\right\}
\end{equation}
where ${\Psi}_1=\psi_{min}$ and $\Psi_{p+1}=\psi_{max}$, $\phi_{min}$, $\phi_{max}$, $\psi_{min}$ and $\psi_{max}$ are constant,  $\phi_{min}<\phi_{max}$, and $\psi_{min}<\psi_{max}$. We divide $\mathcal{C}$ into $p$ horizontal sub-rectangles, therefore,
\vspace{-0.3cm}
\begin{equation}
\mathcal{C}=\bigcup_{j=1}^{p}\mathcal{C}_j,
\end{equation}
where $\mathcal{C}_j$ defines the $j$-th sub-rectangle in planning space. We distribute a uniform $m_\phi\times m_j$ grid over $\mathcal{C}_j$.
Therefore, a total of $m_\phi\cdot m_\psi $ of nodes are distributed over the planing space, where $m_\psi=m_1+\cdots+m_p$. We use $\partial \mathcal{C}_{j,1}$, $\partial \mathcal{C}_{j,2}$, $\partial \mathcal{C}_{j,3}$, and $\partial \mathcal{C}_{j,4}$ to define the bottom, right, top, and left boundaries of $\mathcal{C}_j$, respectively. We note that the $m$ nodes scattered over $\partial \mathcal{C}_{j,1}$ coincide with the $m$ nodes distributed over  $\partial \mathcal{C}_{j,3}$.
\begin{algorithm}
  \caption{Distribution of $n_{j,b}$ nodes over boundary $\partial \mathcal{C}_{j,b}$, for $j=1,\cdots,p$ and $b=1,\cdots,4$.}\label{Alg_pointdist}
  \begin{algorithmic}[1]
          \State \textit{Get:} Line segment end points along $\partial \mathcal{C}_{j,b}$ ($\begin{bmatrix}
    \bar{x}_{h,j,b}&
    \bar{y}_{h,j,b}
\end{bmatrix}^T$ for every $h=1,\cdots,\gamma_{j,b}$); total number of nodes along ($\partial \mathcal{C}_{j,b}$  $n_{j,b}$)
         \State \textit{Obtain:} Positions of the nodes distributed over $\partial \mathcal{C}_{j,b}$.
         \State $L_{j,b}\leftarrow 0$.
        \For{\texttt{$h\in \left\{1,\cdots,\gamma_{j,b}\right\}$}}
            \State Compute $l_{h,j,b}$ using Eq. \eqref{lhjb}.
            \State $L_{j,b}\leftarrow L_{j,b}+l_{h,j,b}$
        \EndFor
        \State $\Delta L_{j,b}={L_{j,b}\over n_{j,b}-1}$.
        \For{\texttt{$h\in \left\{1,\cdots,\gamma_{j,b}\right\}$}}
            \For{\texttt{$i\in \left\{1,\cdots,n_{j,b}\right\}$}}
                \State Compute $\mathbf{\Omega}_{i,h,j,b}$ using Eq. \eqref{omegaijb}.
                \If{$\mathbf{\Omega}_{i,h,j,b}\geq \mathbf{0}$}
                    \State Compute $x_i$ and $y_i$ using Eq. \eqref{xiyi}.
                \EndIf
            \EndFor
        \EndFor
  \end{algorithmic}
\end{algorithm}

\subsection{Boundary Conditions}\label{Boundary Conditions}
To obtain boundary conditions of PDEs \eqref{xpp} and \eqref{ypp}, the nodes on $\partial \mathcal{C}_{j,b}$ are uniformly mapped over $\partial \mathcal{S}_{j,b}$, for   $j=1,\cdots,p$  and $b=1,\cdots,4$. Then, every node $i$, positioned at $\left(\phi_i,\psi_i\right)\in \partial\mathcal{C}_{j,b}$ has corresponding position $\left(x_i,y_i\right)\in \partial \mathcal{S}_{j,b}$. Therefore, $x_i=x\left(\phi_i,\psi_i\right)$ and  $y_i=y\left(\phi_i,\psi_i\right)$ are known along the boundary $\partial \mathcal{C}_{j}$, and we can solve PDEs \eqref{xpp} and \eqref{ypp} as a Dirichlet problem.
\begin{figure}[htbp]
\centering
\includegraphics[width=3.3in]{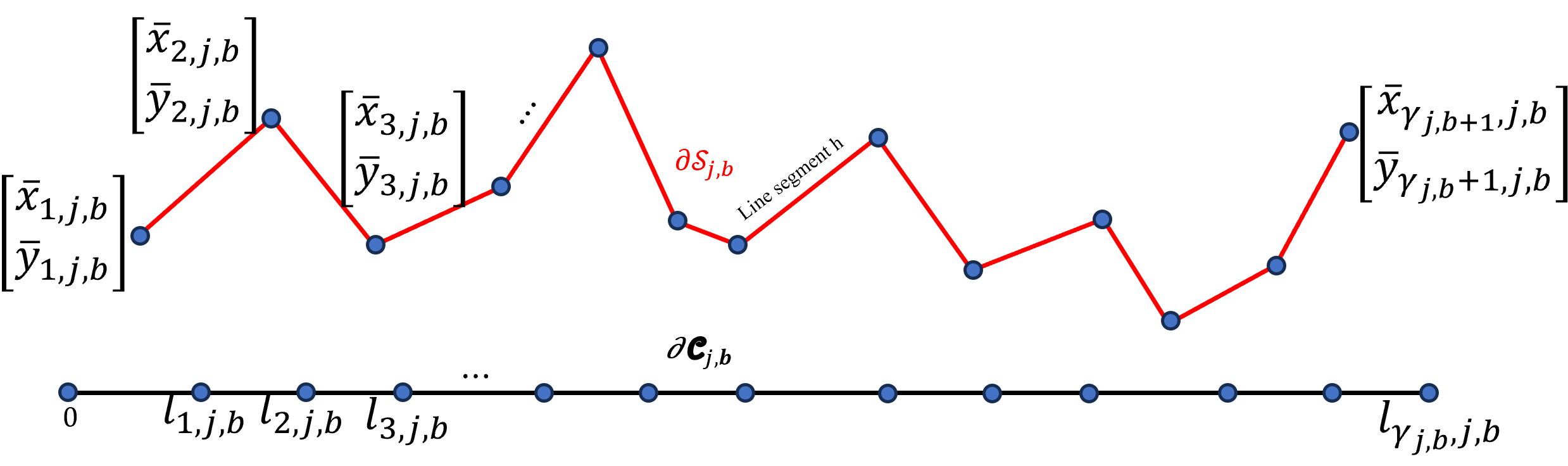}
\caption{Mapping between $\partial \mathcal{S}_{j,b}$ and $\partial \mathcal{C}_{j,b}$ ($j=1,\cdots, p$ and $b=1,\cdots,4$).}
\label{BoundaryMapping}
\end{figure}
We use Algorithm \ref{Alg_pointdist} to uniformly distribute $n_{j,b}$ nodes along $\partial \mathcal{C}_{j,b}$,  when $\partial \mathcal{C}_{j,b}$ consists $\gamma_{j,b}$ serially connected line segments connecting $\gamma_{j,b}+1$ points given by finite sequence 
\vspace{-0.25cm}
\[
\partial \mathcal{C}_{j,b}:~\begin{bmatrix}
    \bar{x}_{1,j,b}\\
    \bar{y}_{1,j,b}\\
\end{bmatrix}
\cdots
\begin{bmatrix}
    \bar{x}_{\gamma_{j,b}+1,j,b}\\
    \bar{y}_{\gamma_{j,b}+1,j,b}\\
\end{bmatrix}
,\qquad j=1,\cdots,p,~b=1,\cdots,4.
\]
We define
\vspace{-0.25cm}
\begin{subequations}
    \begin{equation}\label{lhjb}
        l_{h,j,b}=\sqrt{\left(\bar{x}_{h+1,j,b}-
    \bar{x}_{h+1,j,b}\right)^2+\left(\bar{y}_{h+1,j,b}-
    \bar{y}_{h+1,j,b}\right)^2}
      \end{equation}
      \vspace{-0.25cm}
        \begin{equation}
        L_{j,b}=\sum_{h=1}^{\gamma_{j,b}}l_{h,j,b}
      \end{equation}
      \vspace{-0.25cm}
              \begin{equation}
        \Delta L_{j,b}= {L_{j,b}\over n_{j,b}-1}
      \end{equation}
\end{subequations}
as the length of the $h$-th line segment on $\partial \mathcal{C}_{j,b}$, the total length of boundary  $\partial \mathcal{C}_{j,b}$, and length increment, respectively, for $j=1,\cdots,p$, $h=0,\cdots,\gamma_{j,b}-1$, and $b=1,\cdots,4$ {(See Fig. \ref{BoundaryMapping})}. 
\vspace{-0.25cm}
\begin{definition}\label{Def1}
Given boundary node $i$, positioned at $(x_i,y_i)\in \partial \mathcal{C}_{j,b}$, and end points of the $h$-line segments of $\mathcal{C}_{j,b}$ positioned at $\begin{bmatrix}
    \bar{x}_{h,j,b}&
    \bar{y}_{h,j,b}
\end{bmatrix}^T$ and $\begin{bmatrix}
    \bar{x}_{h+1,j,b}&
    \bar{y}_{h+1,j,b}
\end{bmatrix}^T$, we define vector operator
    We say the $i$-th node $(x_i,y_i)\in \partial \mathcal{C}_{j,b}$ is on the $h$-th line segment connecting $\begin{bmatrix}
    \bar{x}_{h,j,b}&
    \bar{y}_{h,j,b}
\end{bmatrix}^T$ and $\begin{bmatrix}
    \bar{x}_{h+1,j,b}&
    \bar{y}_{h+1,j,b}
\end{bmatrix}^T$, we define vector operator
\begin{equation}\label{omegaijb}
\mathbf{\Omega}_{i,h,j,b}=\begin{bmatrix}
    {l}_{h,j,b}&{l}_{h+1,j,b}\\
    1&1
\end{bmatrix}
^{-1}
\begin{bmatrix}
    i\Delta L_{j,b}\\
    1
\end{bmatrix}
\end{equation}
for $i=1,\cdots,n_{j,b}$, $j=1,\cdots,p$, $b=1,2,3,4$, and $h=0,\cdots,\gamma_{j,b}-1$.
\end{definition}
By using Definition \ref{Def1}, we can check whether boundary  node $i$ is located on the $h$-th line segment or not. If $\mathbf{\Omega}_{i,h,j,b}\geq \mathbf{0}$, then, 
\vspace{-0.25cm}
\begin{subequations}\label{xiyi}
    \begin{equation}\label{xi}
        x_i=\mathbf{\Omega}_{i,h,j,b}^T\begin{bmatrix}
    \bar{x}_{h,j,b}&
    \bar{x}_{h+1,j,b}
\end{bmatrix}^T
    \end{equation}\label{yi}
        \begin{equation}
        y_i=\mathbf{\Omega}_{i,h,j,b}^T\begin{bmatrix}
    \bar{y}_{h,j,b}&
    \bar{y}_{h+1,j,b}
\end{bmatrix}^T
    \end{equation}
\end{subequations}
assign the $x$ and $y$ components of the $i$-th node $\partial \mathcal{C}_{j,b}$.

\begin{figure}[htbp]
\centering
\includegraphics[width=3.3in]{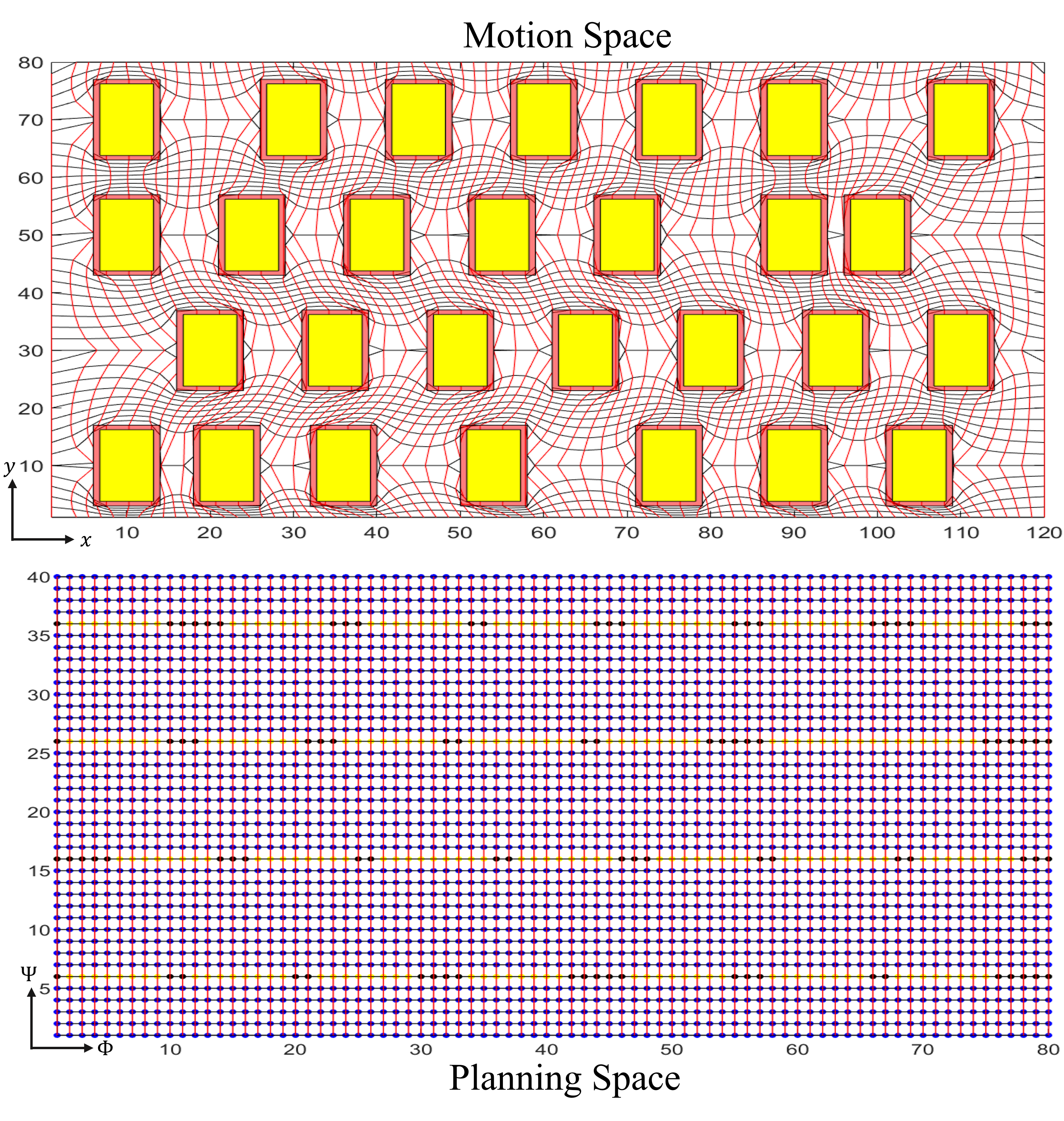}
 \vspace{-0.3cm}
\caption{Top: Motion space. Bottom: Planning space.}
\label{TemporalPlanningPict}
\end{figure}

\section{Temporal Planning}\label{Temporal Planning}
The objective of temporal planning is to obtain the desired trajectory between  from the initial position to target destination in $\mathcal{P}$ minimizing the travel distance. We define this problem as an A* search over the uniform grid distributed in $\mathcal{C}$. This search problem holds the following unique properties:
\begin{enumerate}
    \item \textbf{Property 1:} Operation, evaluation, and heuristic costs all depend on geodesic nodal distances. \label{cond1}
    \item \textbf{Property 2:} There is no obstacle in the planning space, however, transitions over the uniform grid, distributed over the planning space $\mathcal{C}$, constrained over the nodes representing the obstacle boundaries in the actual motion space.\label{cond2}
\end{enumerate}
We note that Property \ref{cond2} constrains paths designed over the planning space to  not cross over the nodes representing obstacle boundaries in the planning space. For a better clarification, the bottom sub-figure in Fig.  \ref{TemporalPlanningPict} is the planning space associated with the obstacle motion space shown in the top sub-figure, in Fig. \ref{TemporalPlanningPict},  where the yellow  dots represent the nodes distributed over the obstacle boundaries. It is not authorized that a robot trajectory, which is assigned by the A* searching in the planning space, crosses the yellow nodes in the planning space. 

We showed that the implementation of an A* search in the planning space $\mathcal{C}$ can generate a shorter path, compared to the existing A* search conducted over the motion space $\mathcal{P}$. This will be discussed in Section \ref{Results}.

\section{Trajectory Tracking Control}\label{Trajectory Tracking Control}
Without loss of generality, this paper assumes that the robot is a quadcopter modeled by the dynamics presented in \cite{rastgoftar2021safe, el2023quadcopter}. By using the feedback liberalization approach (presented in \cite{rastgoftar2021safe, el2023quadcopter}), the quadcopter dynamics are converted to the following external and internal dynamics:
\begin{subequations}
    \begin{equation}\label{EXTDYN}
        \mathbf{x}_{k+1}=\mathbf{A}_p\mathbf{x}_k+\mathbf{B}_p\mathbf{u}_k,\qquad k\in \mathbb{N}
    \end{equation}
   \begin{equation}
        \mathbf{z}_{k+1}=\begin{bmatrix}
            1&\Delta T\\
            0&1
        \end{bmatrix}\mathbf{z}_k+\begin{bmatrix}
            0\\
            1
        \end{bmatrix}\mathbf{v}_k,\qquad k\in \mathbb{N},
    \end{equation}
\end{subequations}
where $k$ denotes the discrete time; $\mathbf{x}_k=\begin{bmatrix}
    \mathbf{r}_k^T&\mathbf{v}_k^T&\mathbf{a}_k^T&\mathbf{j}_k^T
\end{bmatrix}^T\in \mathbb{R}^{12\times 1}$ aggregates quadcopter's position, velocity, acceleration,  and jerk; $\mathbf{z}_k=\begin{bmatrix}
    \psi_k&\dot{\psi}_k
\end{bmatrix}^T\in \mathbb{R}^{2\times 1}$ aggregates yaw and time derivative of yaw;  $\mathbf{u}_k\in \mathbb{R}^{3\times 1}$ is the snap control; and $\mathbf{v}_k\in \mathbb{R}$ is the control input of the internal dynamics. Also, 
\begin{subequations}
    \begin{equation}
        \mathbf{A}_p=\mathbf{I}_{12}+\Delta T\begin{bmatrix}
            \mathbf{0}_{3\times 3}&\mathbf{I}_3&\mathbf{0}_{3\times 3}&\mathbf{0}_{3\times 3}\\
            \mathbf{0}_{3\times 3}&\mathbf{0}_{3\times 3}&\mathbf{I}_3&\mathbf{0}_{3\times 3}\\
            \mathbf{0}_{3\times 3}&\mathbf{0}_{3\times 3}&\mathbf{0}_{3\times 3}&\mathbf{I}_3\\
            \mathbf{0}_{3\times 3}&\mathbf{0}_{3\times 3}&\mathbf{0}_{3\times 3}&\mathbf{0}_{3\times 3}\\
        \end{bmatrix}
        \in \mathbb{R}^{12\times 12},
    \end{equation}
        \begin{equation}
        \mathbf{B}_p=\Delta T\begin{bmatrix}
            \mathbf{0}_{3\times 3}&
            \mathbf{0}_{3\times 3}&
            \mathbf{0}_{3\times 3}&
            \mathbf{I}_{3}
        \end{bmatrix}
        ^T
        \in \mathbb{R}^{12\times 3},
    \end{equation}
\end{subequations}
are time-invariant, where time increment $\Delta T$ is constant, $\mathbf{I}_3\in \mathbb{R}^{3\times 3}$ is the identity matrix, and  $\mathbf{0}_{3\times 3}\in \mathbb{R}^{3\times 3}$ is the zero-entry matrix. We choose
\begin{equation}
    \mathbf{v}_k=\mathbf{k}_\psi\mathbf{z}_k
\end{equation}
and select constant gain matrix $\mathbf{k}_\psi$ such that eigenvalues of the characteristic equation
\begin{equation}
    \left|\zeta \mathbf{I}_2-\mathbf{A}_\psi\right|=0
\end{equation}
are inside the unit disk centered at the origin. Therefore, $\psi_k$ asymptotically converges to $0$, and therefore, the quadcopter motion is modeled by the external dynamics \eqref{EXTDYN}.

\subsection{Safety Constraints}\label{Safety Constraints}
\begin{figure}[htbp]
\centering
\includegraphics[width=3.2 in]{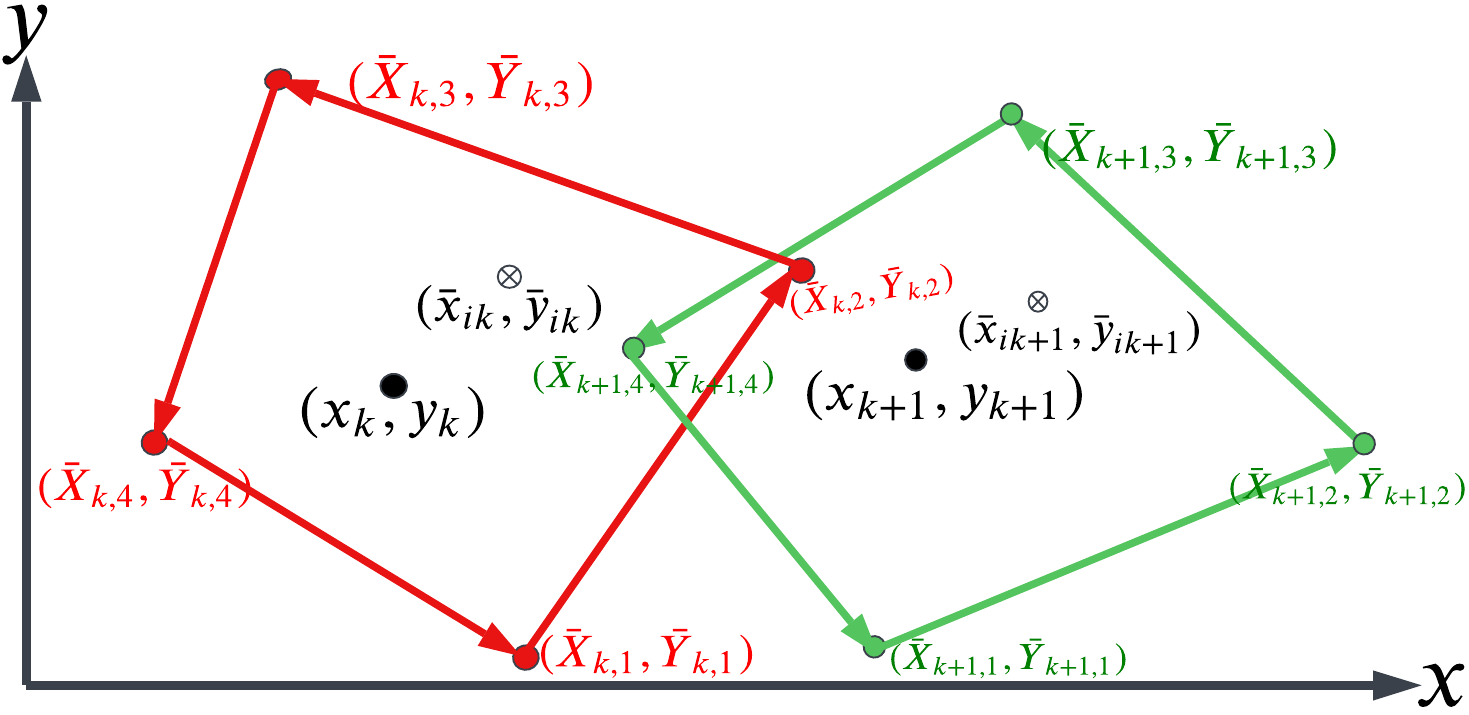}
\caption{Schematic of two consecutive quadrangles containing the robot's desired and actual at discrete times $k$ and $k+1$.}
\vspace{-0.25cm}
\label{NeighboringQuadrangles}
\end{figure}

We suppose that the quadcopter moves horizontally at elevation $\bar{z}_0$ and specify the desired trajectory by sequence
\begin{equation}
    \mathbf{r}_d=\mathbf{p}_1\cdots \mathbf{p}_m,
\end{equation}
where $\mathbf{p}_k=\begin{bmatrix}
    \bar{x}_{i_k}&\bar{y}_{j_k}
\end{bmatrix}^T $ aggregates $x$ and $y$ components of the quadcopter's  desired way-point obtained by the planning approach discussed in Section \ref{BoundaryConditionsSpecification}, through A* searching in $\mathcal{C}$. We note that a nonsingular mapping exists between $\left(i_k,j_k\right)\in \mathcal{C}$ and    $\left(\bar{x}_{i_k},\bar{y}_{j_k}\right)\in \mathcal{S}$. Desired position  $\left(\bar{x}_{i_k},\bar{y}_{j_k}\right)$ is enclosed by a \textit{neighboring quadrangle} whose vertices are positioned at  $\left(\bar{X}_{k,1},\bar{Y}_{k,1}\right)=\left(\bar{x}_{i_k-1},\bar{y}_{j_k-1}\right)$, $\left(\bar{X}_{k,2},\bar{Y}_{k,2}\right)=\left(\bar{x}_{i_k+1},\bar{y}_{j_k-1}\right)$, $\left(\bar{X}_{k,3},\bar{Y}_{k,3}\right)=\left(\bar{x}_{i_k+1},\bar{y}_{j_k+1}\right)$, and  $\left(\bar{X}_{k,4},\bar{Y}_{k,4}\right)=\left(\bar{x}_{i_k-1},\bar{y}_{j_k+1}\right)$. Figure \ref{NeighboringQuadrangles} shows the schematic of the neighboring quadrangle at two consecutive times $k$ and $k+1$. To ensure collision avoidance, $x$ and $y$ components of quadcopter's actual position, denoted by $x_k$ and $y_k$,  are constrained to be inside the neighboring quadrangle at every time $k$. By defining $\bar{X}_{k,5}=\bar{X}_{k,1}$ and $\bar{Y}_{k,5}=\bar{Y}_{k,1}$, this safety requirement is assured, if the following inequality condition is satisfied:
\begin{equation}\label{equationsafety}
\bigwedge_{h=1}^{n_\tau}\bigwedge_{j=1}^{4}\left(\begin{vmatrix}
        x_{{k+h}}-\bar{X}_{k,j}&y_{{k+h}}-\bar{Y}_{k,j}\\
        \bar{X}_{k,j+1}-\bar{X}_{k,j}&\bar{Y}_{k,j+1}-\bar{Y}_{k,j}\\
    \end{vmatrix}
    \leq 0\right).
\end{equation}

We can rewrite Eq. \eqref{equationsafety} as
\vspace{-0.25cm}
\begin{equation}\label{mainsafety}
    \bigwedge_{h=1}^{n_\tau}\left(\mathbf{\Lambda}_{i_{k+h}}\begin{bmatrix}
        x_{{k+h}}\\
        y_{{k+h}}
    \end{bmatrix}\leq \mathbf{\Gamma}_{i_{k+h}}\right),
\end{equation}
where 
\vspace{-0.25cm}
\begin{subequations}
    \begin{equation}
    \mathbf{\Lambda}_{i_{k+h}}=\begin{bmatrix}
        \bar{Y}_{k,2}-\bar{Y}_{k,1}&\bar{X}_{k,1}-\bar{X}_{k,2}\\
        \vdots&\vdots\\
         \bar{Y}_{k,5}-\bar{Y}_{k,4}&\bar{X}_{k,4}-\bar{X}_{k,5}\\
    \end{bmatrix}
    ,
    \end{equation}
    \begin{equation}
    \mathbf{\Gamma}_{i_{k+h}}=\begin{bmatrix}
        \bar{X}_{k,1}\left(\bar{Y}_{k,2}-\bar{Y}_{k,1}\right)-\bar{Y}_{k,1}\left(\bar{X}_{k,2}-\bar{X}_{k,1}\right)\\
        \vdots\\
        \bar{X}_{k,4}\left(\bar{Y}_{k,5}-\bar{Y}_{k,4}\right)-\bar{Y}_{k,4}\left(\bar{X}_{k,5}-\bar{X}_{k,4}\right)
    \end{bmatrix},
    \end{equation}    
\end{subequations}

\subsection{MPC-Based Trajectory Tracking}\label{MPC-Based Trajectory Tracking}
We use the MPC method to determine the control input $\mathbf{u}_k$ so that the $\mathbf{p}_k$ is stably tracked while the safety constraints given by Eq. \eqref{mainsafety} are  satisfied. To this end, we define 
\vspace{-0.25cm}
\begin{equation}
\resizebox{0.99\hsize}{!}{%
 $
    J_k={1\over 2}\sum_{h=0}^{n_\tau-1}\left[\mathbf{u}_{k+h}^T\mathbf{u}_{k+h}+\beta\left(\begin{bmatrix}
        x_{k+h+1}\\y_{k+h+1}
    \end{bmatrix}-\mathbf{p}_{k+h+1}\right)^T\mathbf{f}_h\left(\begin{bmatrix}
        x_{k+h}\\y_{k+h}
    \end{bmatrix}-\mathbf{p}_{k+h+1}\right) \right]
    $
    }
\end{equation}
as the MPC cost, where $\mathbf{f}_h$ is a diagonal weight matrix, $n_\tau $ is the length of the prediction window, $x_k$ and $y_k$ denote quadcopter's actual position components at time $k$,  and $\beta>0$ is a scaling factor. 

Given dynamics \eqref{EXTDYN}, we define vector $\mathbf{Y}_k=\begin{bmatrix}\mathbf{x}_{k+1}^T&\cdots&\mathbf{x}_{k+n_\tau}^T\end{bmatrix}$ and relate it to $\mathbf{x}_k$ (the state vector at time $k$) and $\mathbf{U}_k =\begin{bmatrix}\mathbf{u}_{k+1}^T&\cdots&\mathbf{x}_{k+n_\tau-1}^T\end{bmatrix}$ by
\begin{equation}
    \mathbf{Y}_k=\mathbf{G}+\mathbf{H}\mathbf{U}_k,
\end{equation}
where $\mathbf{G}=\left[\mathbf{G}_{i}\right]\in \mathbb{R}^{12n_\tau\times 12}$ and $\mathbf{H}=\left[\mathbf{H}_{i,j}\right]\in \mathbb{R}^{12n_\tau\times 3n_\tau}$ are obtained as follows:
\vspace{-0.25cm}
\begin{subequations}
    \begin{equation}
    \mathbf{G}_{i}=\mathbf{A}_p^i,
    \end{equation}
    \begin{equation}
        \mathbf{H}_{i,j}=
        \begin{cases}
             \mathbf{A}_p^{i-j}\mathbf{B}_p&j<i\\
             \mathbf{0}&\mathrm{else}
        \end{cases} 
        .
    \end{equation}
\end{subequations}
We also use $\mathbf{R}_k=\begin{bmatrix}
    x_{k+1}&y_{k+1}&\cdots&x_{k+n_\tau}&y_{k+n_\tau}
\end{bmatrix}^T$ to aggregate the prediction of $x$ and $y$ components of the quadcopter's actual position at time $k+1$ through $k+n_\tau$, where $\mathbf{R}_k$ and $\mathbf{Y}_k$ are related by
\vspace{-0.25cm}
\begin{equation}
    \mathbf{R}_k=\mathbf{C}_p\mathbf{Y}_k
\end{equation}
with $\mathbf{C}_p=\mathbf{I}_{n_\tau}\otimes \begin{bmatrix}
    \mathbf{I}_2&\mathbf{0}_{2\times 10}
\end{bmatrix}$. Furthermore, we use $\mathbf{P}_k=\begin{bmatrix}
    \mathbf{p}_{k+1}^T&\cdots&\mathbf{p}_{k+n_\tau}^T
\end{bmatrix}^T$ to aggregate the robot desired positions at time steps $k+1$ through $k+n_\tau$. 

The control objective is to minimize $J_k$ such that safety constraint equations, given by Eq. \eqref{mainsafety}, is satisfied, where $z_k$ is enforced to be equal to $\bar{z}_0$ at every discrete time $k$. This problem can be formulated as a quadratic programming by
\[
\min {1\over 2}\mathbf{U}_k\mathbf{W}_1\mathbf{U}_k+\mathbf{w}_2\mathbf{U}_k
\]
subject to
\[
\mathbf{A}_k^{ineq}\mathbf{U}_k\leq \mathbf{b}_k^{ineq},
\]
\[
\mathbf{A}_k^{eq}\mathbf{U}_k=\mathbf{b}_k^{eq},
\]
where
\vspace{-0.25cm}
    \begin{equation}
        \mathbf{W}_1=\mathbf{I}_{3n_\tau}+\beta \mathbf{H}^T\mathbf{C}_p^T\mathbf{diag}\left(\mathbf{f}_0,\cdots, \mathbf{f}_{n_\tau-1}\right)\mathbf{C}_p\mathbf{H},
    \end{equation}
    \begin{equation}
        \mathbf{w}_2=\beta \left( \mathbf{C}_p\mathbf{G}\mathbf{x}_k-\mathbf{P}_k\right)^T\mathbf{C}_p\mathbf{H},
    \end{equation}
    \begin{equation}
    \mathbf{A}_k^{ineq}=\mathbf{diag}\left(
            \mathbf{\Lambda}_{i_{k+1}},\cdots,\mathbf{\Lambda}_{i_{k+n_\tau}}        
        \right)\mathbf{C}_p\mathbf{H},
    \end{equation}
    \begin{equation}
      \begin{split}
        \mathbf{b}_k^{ineq}=&\begin{bmatrix}
            \mathbf{\Gamma}_{i_{k+1}}^T&\cdots&\mathbf{\Gamma}_{i_{k+n_\tau}}^T         
        \end{bmatrix}^T\\
        -&\mathbf{diag}\left(
            \mathbf{\Lambda}_{i_{k+1}},\cdots,\mathbf{\Lambda}_{i_{k+n_\tau}}        
        \right)\mathbf{C}_p\mathbf{G}\mathbf{x}_k,
        \end{split}
    \end{equation}
        \begin{equation}
     \mathbf{A}_k^{eq}= \left(\mathbf{I}_{n_\tau}\otimes\begin{bmatrix}\mathbf{0}_{1\times 2}&1&\mathbf{0}_{1\times 9}\end{bmatrix}\right)\mathbf{H},
    \end{equation}
    \begin{equation}
     \mathbf{b}_k^{eq}=\bar{z}_0\mathbf{1}_{n_\tau\times 1}-\left(\mathbf{I}_{n_\tau}\otimes\begin{bmatrix}\mathbf{0}_{1\times 2}&1&\mathbf{0}_{1\times 9}\end{bmatrix}\right)\mathbf{G}\mathbf{x}_k.
    \end{equation}


\section{RESULTS}\label{Results}
We consider quadcopter motion planning and control in the obstacle-laden motion space shown in Fig. \ref{TemporalPlanningPict}. We use the spatial planning method presented in Section \ref{Spatial Planning} to split the motion space into navigable space $\mathcal{S}$ and obstacle zone  $R$, where: $\mathcal{S}$ is decomposed into $p=5$ navigable channels $\mathcal{S}_1$ through $\mathcal{S}_5$; $\mathcal{S}_j$ and $\mathcal{S}_{j+1}$ sandwich all obstacles defined by $\mathcal{R}_j$, for $j=1,\cdots,p-1$, as illustrated in Fig.  \ref{ExampleMotionSpace}. We use the algorithmic approach presented in Section \ref{Boundary Conditions} to solve the governing PDE \eqref{ellipticpde} using the Finite Difference method. We obtain the potential and steam lines  that are  shown by red and black curves in Fig.   \ref{TemporalPlanningPict}.

\begin{figure}[h]
\centering
\includegraphics[width=3.3in]{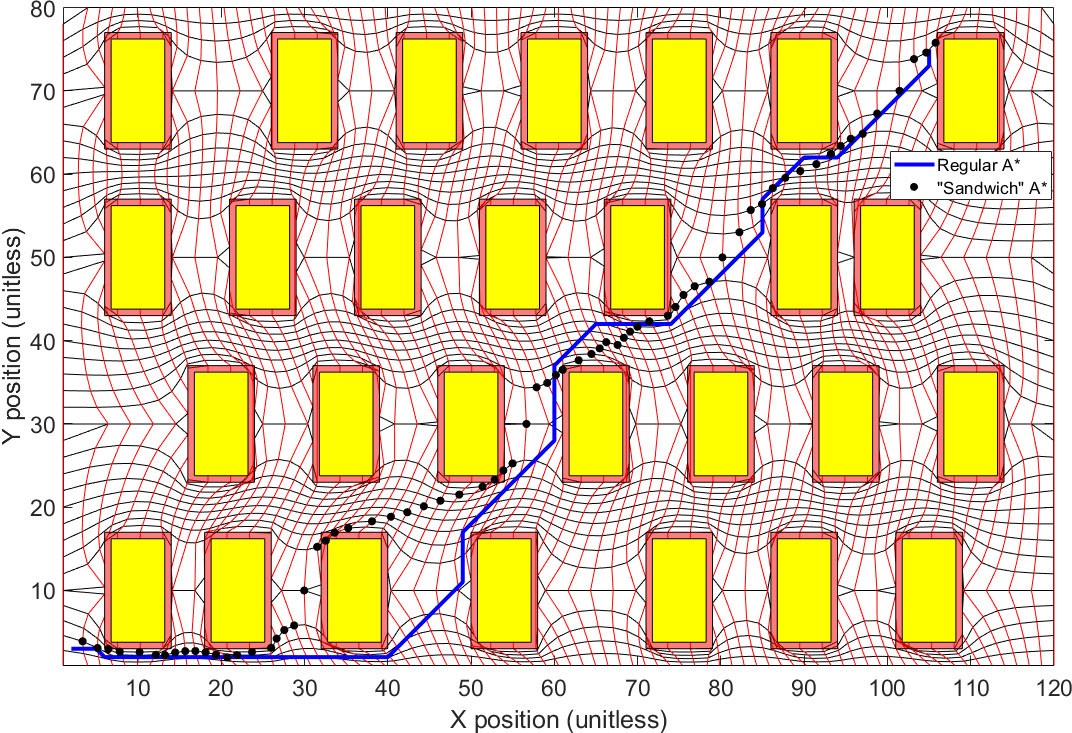}
\vspace{-0.25cm}
\caption{``Sandwich'' and regular A* paths.}
\label{fig:MPC}
\end{figure}

\begin{figure}[h]
\centering
\includegraphics[width=3.3in]{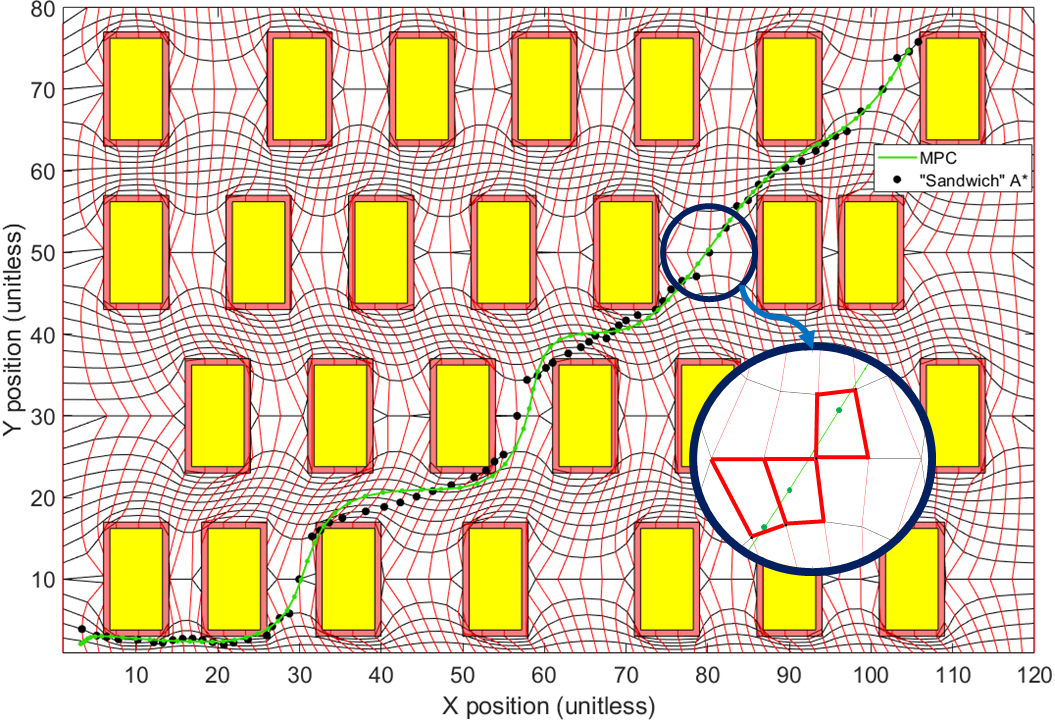}
\vspace{-0.25cm}
\caption{Quadcopter's actual path shown by green. }
\label{trajectorytracking}
\end{figure}

\begin{figure}[htb]
\centering
\includegraphics[width=3.3in]{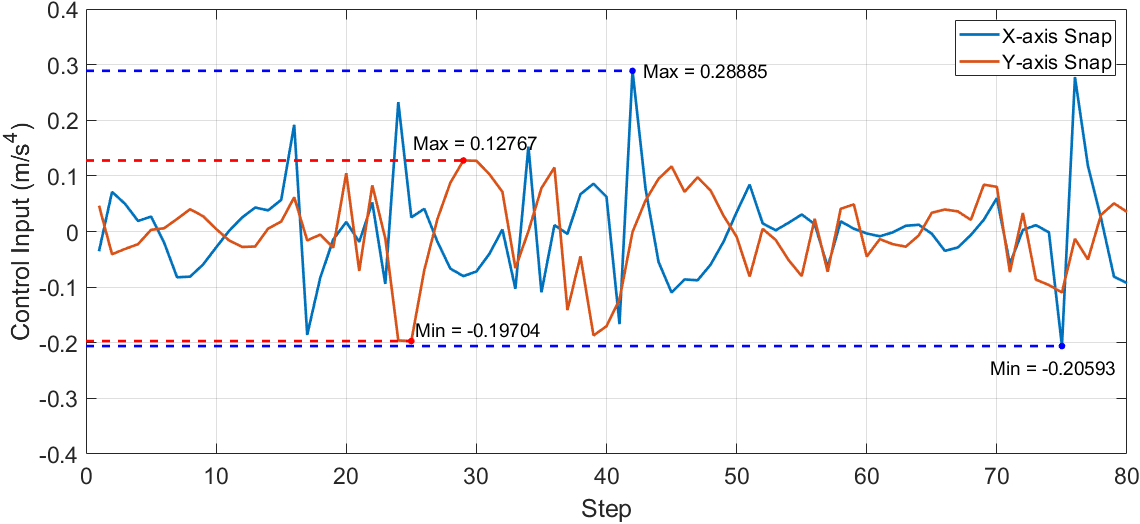}
\vspace{-0.25cm}
\caption{Quadcopter's control input $\mathbf{u}_k$ versus discrete time $k$.}
\label{fig:MPC_snap}
\end{figure}

 Given initial and final positions, we applied A* to obtain the quadcopter desired trajectory by searching over the planning space as detailed in Section \ref{Temporal Planning}. This desired trajectory is shown by black dots in Fig. \ref{fig:MPC}. By applying regular A*, through searching over the motion space, we also obtained the blue path shown in Fig.  \ref{fig:MPC} as the optimal path. Comparing the results, we observe $6.21\%$ of length reduction when the A* search is conducted over the planning space (See Table \ref{table:compastars}). By  adopting the quadcopter model presented in Section \eqref{PROBLEM STATEMENT}, we applied the MPC-based trajectory tracking control, presented in Section  \ref{MPC-Based Trajectory Tracking}, to stably and safely track the quadcopter's desired trajectory where all safety requirements. The quadcopter's actual path   is shown by green \ref{trajectorytracking}. Figure \ref{trajectorytracking} shows that the quadcopters' actual positions are enclosed by the neighboring quadrangles. Figure  \ref{fig:MPC_snap} plots the $x$ and $y$ components of the snap control $\mathbf{u}_k$ versus discrete time $k$.
\begin{table}[h!]
\centering
\caption{Path lengths of ``Sandwich'' and regular  A* search.}
\label{table:compastars}
\begin{tabular}{|c | c|c|} 
 \hline
& Existing A* &Sandwich A* \\ 
 \hline
 Length& $145.95m$ & $138.51m$ \\ 
 \hline
\end{tabular}
\end{table}




\section{CONCLUSION}\label{Conclusion}
We developed a new approach for motion planning in constrained environments. We showed how we can exclude obstacles from the motion space by establishing a nonsingular mapping between the motion space and the planning space. We also obtained shorter paths between initial and target locations by applying search methods over the planning space instead of searching over the motion space. We also developed an MPC-based trajectory tracking control for a quadcopter navigating in an obstacle-laden environment, where we abstained and enforced safety requirements as linear constraints.

\section*{Acknowledgment}
This work has been supported by the National Science Foundation under Award Nos. 2133690 and 1914581.
\bibliographystyle{IEEEtran}
\bibliography{refs}
\end{document}